\documentclass[default,longbib]{aastex7}
\usepackage{varwidth}
\usepackage{amsmath, amsthm}
\usepackage{wrapfig,booktabs,lipsum}
\usepackage{algorithm}
\usepackage{algpseudocode}
\usepackage{natbib}

\newcommand{\forjournal}[1]{}
\newcommand{\tc}[2]{\textcolor{#1}{#2}}

\newcommand{\reals}{{\mathbb{R}}}

\newcommand{\norm}[1]{\left\lVert#1\right\rVert}
\newcommand{\mnorm}[1]{{\left\vert\kern-0.25ex\left\vert\kern-0.25ex\left\vert #1 
    \right\vert\kern-0.25ex\right\vert\kern-0.25ex\right\vert}}

\newcommand{\mc}{\mathcal}

\newcommand{\G}{\mathcal{G}}
\newcommand{\V}{\mathcal{V}}
\newcommand{\E}{\mathcal{E}}
\newcommand{\A}{\mathcal{A}}
\newcommand{\bB}{\mathbf{\mathcal{B}}}

\newtheorem{theorem}{Theorem}
\newtheorem{lemma}{Lemma}

\newtheorem{proposition}{Proposition}

\begin{document}
\title{A Markov Decision Process Framework for Early Maneuver Decisions in Satellite Collision Avoidance}

\author[orcid=0000-0000-0000-0001,sname='North America']{Francesca Ferrara}
\altaffiliation{Equal contributions}
\affiliation{Automatic Control Laboratory, ETH Z{\"u}rich}
\email[show]{fferrara@student.ethz.ch}

\author[orcid=0000-0000-0000-0002,sname='North America']{Lander W. Schillinger Arana}
\altaffiliation{Equal contributions}
\affiliation{C3U Laboratory, Georgia Institute of Technology}
\email[show]{larana9@gatech.edu}

\author[orcid=0000-0000-0000-0003,gname=Bosque, sname='Sur America']{Florian D{\"o}rfler} 
\altaffiliation{}
\affiliation{Automatic Control Laboratory, ETH Z{\"u}rich}
\email[show]{doerfler@control.ee.ethz.ch}

\author[orcid=0000-0000-0000-0004,gname=Bosque, sname='Sur America']{Sarah H.Q. Li} 
\altaffiliation{}
\affiliation{C3U Laboratory, Georgia Institute of Technology}
\email[show]{sarahli@gatech.edu}







\begin{abstract}
We develop a Markov decision process (MDP) framework to autonomously make guidance decisions for satellite collision avoidance maneuver (CAM) and a reinforcement learning policy gradient (RL-PG) algorithm to enable direct optimization of guidance policy using historic CAM data. In addition to maintaining acceptable collision risks, this approach seeks to minimize the average propellant consumption of CAMs by making \emph{early} maneuver decisions. We model CAM as a continuous state, discrete action and finite horizon MDP, where the critical decision is determining \emph{when} to initiate the maneuver. The MDP models decision rewards using analytical models of collision risk, propellant consumption, and transit orbit geometry. 
By deciding to maneuver earlier than conventional methods, the Markov policy effectively favors CAMs that achieve comparable rates of collision risk reduction while consuming less propellant.  
Using historical data of tracked conjunction events, we verify this framework and conduct an extensive parameter-sensitivity study. When evaluated on synthetic conjunction events, the trained policy consumes significantly less propellant overall and per maneuver in comparison to a conventional cut-off policy that initiates maneuvers $24$ hours before the time of closest approach (TCA). On historical conjunction events, the trained policy consumes more propellant overall but consumes less propellant per maneuver. For both historical and synthetic conjunction events, the trained policy is slightly more conservative in identifying conjunctions events that warrant CAMs in comparison to cutoff policies.
\end{abstract}



\section{Introduction} 
With a growing satellite population in low Earth orbit (LEO), collision risks have become a critical concern for space traffic management~\citep{lal2018global}. Current operational practice~\citep{2023NASAIi} follows a multi-layered process: tracked objects are screened for conjunction events using Two-Line-Elements (TLEs) and orbital propagation tools over a $7$–$10$ day horizon. Potential encounters are flagged and sent to mission operators, who then assess the collision risk in greater detail and explore maneuver options. The decision to execute a CAM is often delayed to improve the reliability of collision risk indicators~\citep{hejduk2019satellite}. While this strategy minimizes unnecessary maneuvers, it often reduces maneuver efficiency---executing maneuvers earlier can \emph{exponentially} decrease propellant consumption by allowing longer coasting durations in the transit orbit~\citep{DeVittori2022Low-ThrustOptimization}, albeit with a higher risk of unnecessarily maneuvering when no collision materializes. 

In this manuscript, we explore whether maneuvering earlier can consume less propellant while minimizing unnecessary CAMs. Specifically, we propose a stochastic decision-making model that uses conjunction data to \emph{anticipate} the average propellant consumption and weigh it against the collision risk. Applying this model to historical conjunction data messages (CDMs), we train a CAM guidance policy that recommends \emph{earlier maneuvers} for conjunctions characterized by higher expected propellant consumption or more reliable collision risk estimates, instead of delaying the maneuver until a conventional cutoff time. 
Our approach provides a systematic way to combine analytical collision risk prediction with reinforcement learning to mitigate collision risks.


\textbf{Contributions}. In the absence of CAM guidance models for early maneuver decision-making, we propose a continuous state, discrete action, finite-horizon MDP. The MDP trades off collision risk reliability with propellant consumption by changing the maneuver time.  Using RL-PG, we train an trained policy to minimize propellant consumption while maintaining safe collision risk levels. We validate the model's performance  using synthetic and historical CDM data sets. We investigate the sensitivity of the trained policy to variations in the MDP model parameters to find configurations that produce trained and stable performance. When compared with a cut-off policy based on current risk mitigation practice~\citep{2023NASAIi}, the trained policy produces similar detection rates of high risk collisions, and similar rates of actions taken under high and low risk conjunctions. We also find that the trained policy's performance is stable and optimal over a large region of parameter configurations. 

\textbf{Assumptions and Limitations}. 
We highlight some simplifying assumptions made in this MDP framework.
For the satellite anticipating a CAM, we assume that it operates only in circular LEO~\citep{bate2020fundamentals}, that it can move away from its operational orbit for extended periods of time~\citep{sanchez2017reliability}, and that it only carries out in-track burns involving prograde CAMs~\citep{Stoll2011TheManagement}. Together, these assumptions capture the most common configuration for LEO conjunction events as well as the preferable maneuver type for CAM. 

\section{Literature Review}
A collision risk mitigation operation in orbit consists of three phases: conjunction assessment~\citep{flohrer2008assessment,clifton2022optimization}, probability of collision estimation~\citep{Foster1992AVehicles}, and maneuver planning~\citep{mueller2008collision,bombardelli2014collision}. Conjunction assessments are typically performed by the US Space Surveillance Network over a catalog of space objects and their positions~\citep{NASA_SST_SoA_IDTracking_2024}, which is maintained using a combination of ground-based radar and optical sensors. Recent efforts in Space Situational Awareness has pushed for additional sensing tools from third-party providers such as Leolabs~\citep{LeoLabs2025} and in-orbit sensing~\citep{ender2011radar,lal2018global,skinner2020small}. For a high risk conjunction, CDMs containing detailed and updated positions are issued~\citep{moomey2020trending,moomey2023trending} and probability of collisions are computed via statistical methods~\citep{akella2000probability,Foster1992AVehicles,alfano2018probability}. Although computable, collision probabilities are often unreliable due to limited availability of observations and the high epistemic uncertainty surrounding the conjunction event~\citep{alfriend1999probability,hejduk2019satellite,balch2019satellite}. Currently maneuver decisions are manually made by satellite operators. Automating the decision-making step is the focus of this manuscript. If a maneuver deemed necessary, multiple factors (e.g. space object catalogs, thruster mechanism, orbital conditions) are consulted to design a CAM~\citep{gonzalo2021analytical,de2022low,morselli2014collision,pavanello2024recursive} with Monte Carlo verification. This step is computationally-involved~\citep{slater2006collision,jochim2011fuel,king_small_2008}. Recently the maneuver design via optimization techniques has been explored~\citep{armellin2021collision,kelly2005probability}.

Artificial intelligence and machine learning are promising approaches to automate CAM operations. Multiple machine learning techniques have been applied to collision avoidance detection~\citep{Uriot2022SpacecraftCompetition,gonzalo2021board,acciarini2021kessler}. 
Automated maneuver guidance via RL has been explored  in~\citep{Temizer2010CollisionProcesses,Kazemi2024SatelliteOptimization,mu2024autonomous}. Recently, a similar approach to modeling CAM guidance via MDPs is taken in~\citep{kuhl2025markov} using synthetic CDMs and Q-learning. To the best of our knowledge, this manuscript is the first to validate MDP-based CAM performance using historical conjunction event observations. 
\section{Problem Statement and Formulation}\label{sec:problem_statement}
\begin{figure}[ht!]
    \centering    \includegraphics[width=\linewidth]{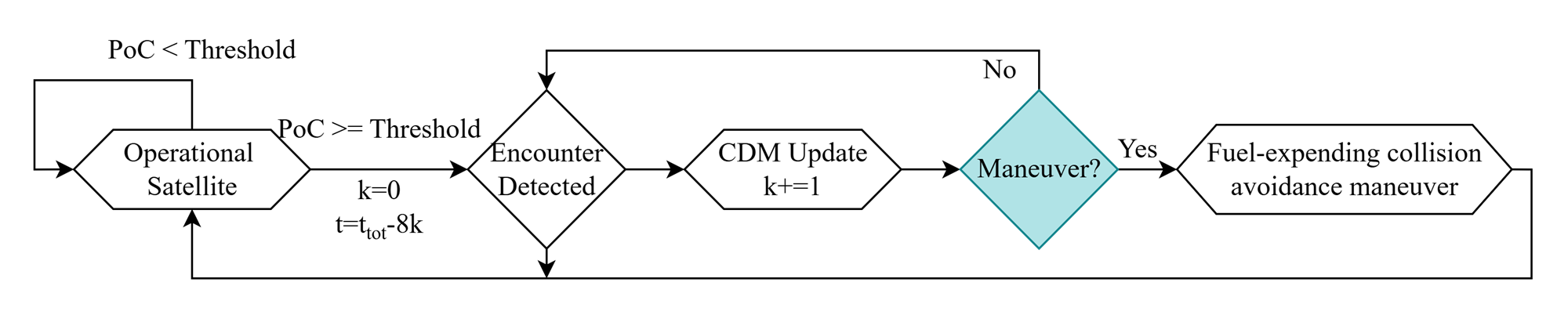}
    \caption{The flow chart representing a CAM process from collision encounter detection to maneuver execution for an operational satellite. This project focuses on automating the maneuver decision-making process, highlighted in blue.}
    \label{fig:cam_process}
\end{figure}

We first provide a brief description of the events leading up to a CAM, also outlined in Fig.~\ref{fig:cam_process}. Consider an operational satellite in LEO with propulsive capabilities. The US Space Surveillance Network constantly monitors potential conjunctions between this satellite and other space objects in their catalog~\citep{NASA_SST_SoA_IDTracking_2024}. The probability of collision (PoC) between the satellite and the debris is evaluated for each potential collision. While the PoC is within a certain threshold, the operational satellite is considered ``safe" and the encounter remains ``undetected". However, if the PoC exceeds this threshold, the Surveillance Network initiates a TCA countdown and periodic CDM updates. In approximately eight hour intervals, the satellite under collision risk receives collision estimation parameter updates via CDMs. Under NASA's recommended procedures, operators decide whether to perform a CAM 24 hours before TCA. For this work, only the satellite can be maneuvered. 
As shown in Sec.~\ref{sec:CAM} and~\ref{sec:FUEL}, earlier executions of CAMs can consume less propellant while still producing the same PoC reduction. However, collision risk indicators from early CDMs do not reliably predict the true collision risk~\citep{Hejduk2019SatelliteApproaches,Balch2016AAnalysis}. We explore whether earlier maneuvers can be executed reliably using the most recent CDM, and explore whether decision-making models such as MDPs can autonomously make earlier maneuver decisions for high risk collisions. 

\textbf{Problem Statement}. For an operational satellite receiving CDMs for a conjunction event, can MDPs model the maneuver decision process using real-time CDM data, and can Markov policies maintain critical collision safety standards while minimizing propellant consumption?

\subsection{Conjunction Data Messages (CDMs)}\label{sec:cdms}
A conjunction event represents a potential collision between two orbital objects, and is predicted by a time series of CDMs containing the objects' most recently updated relative dynamical data~\citep{2023NASAIi}. Each event has a required minimum lead time for performing CAMs, which we denote as the cut-off time and standardize to 24 hours before the TCA, the anticipated time of smallest separation between the satellite and debris (i.e. time of collision). 

A CDM provides updated information related to the conjunction including but not limited to the operational satellite's service orbit altitude, its relative position to the debris, the positional uncertainty covariance of both the satellite and the debris, hard-body radius, and so on. Other input parameters that do not reside within the CDM include the satellite's mass, and specific impulse of its propulsion system. We assume that these parameters are known.

Specifically, each CDM contains the most recent estimations of the following relevant parameters:
\begin{enumerate}
     \item $t \in \reals_+$: the time remaining until TCA;
     \item $ R_{sat}, R_{deb} \in \reals_+$: the target satellite and chaser debris hard body radii, respectively;
     \item $ R_{serv} \in \reals_+$: the satellite's service orbit radius;
     \item ${\boldsymbol{\boldsymbol{\rho}}_{sat}}, {\boldsymbol{\boldsymbol{\rho}}_{deb}} \in \reals^3$: the target satellite's and chasing debris's  position vectors in their respective radial-tangential-normal (RTN) frames;
     \item $\boldsymbol{\rho}_r = \boldsymbol{\rho}_{deb} - \boldsymbol{\rho}_{sat}$: the relative position vector of the debris with respect to the satellite body expressed in the RTN frame;
     \item ${\boldsymbol{v}_{sat}}, {\boldsymbol{v}_{deb}}\in \reals^3$: the target satellite's and chasing debris's velocity vectors in their respective RTN frames;
    \item $\rho_r=\norm{\boldsymbol{\rho}_r} \in \mathbb{R}_+$: the miss distance between the target satellite and the chasing debris at TCA;
    \item $\boldsymbol{\Sigma}_{sat}, \boldsymbol{\Sigma}_{deb} \in \reals^{3\times 3}$: the target satellite's and chasing debris's position covariance matrices in their respective RTN frames.
\end{enumerate}
For the short-term encounters considered in this paper, the hyper-kinetic conjunction scenario permits the following simplifying assumptions~\citep{li_review_2022}:
\begin{enumerate}
    \item The relative motion between two objects in the encounter region is rectilinear;
    \item The relative speed $\boldsymbol{v}_r = \boldsymbol{v}_{deb} - \boldsymbol{v}_{sat}$ between the two objects is constant;
    \item The velocity covariances for both the satellite and the debris are negligible relative to their position covariances;
    \item The position uncertainty remains stable during the encounter and can be described by uncorrelated and constant covariance matrices;
    \item Both satellite and debris's positions are modeled as 3D Gaussian random variables  with probability density functions (pdfs) given by
    \begin{equation}\label{eqn:3DPDF}
        f(\boldsymbol{\rho}_r,\boldsymbol{\Sigma}_i) = \frac{1}{\sqrt{(2\pi)^3\det(\boldsymbol{\Sigma}_i)}}\exp\left[-\frac{1}{2}\boldsymbol{\rho}_r^\top\boldsymbol{\Sigma}_i^{-1}\boldsymbol{\rho}_r\right],\quad\forall i\in\{sat, deb\}.
    \end{equation}
    
\end{enumerate}
By approximating the satellite and debris by their circumscribing spheres with radii $R_{sat}$ and $R_{deb}$ respectively, the Hard Body Radius (HBR) is defined as
\begin{equation}\label{eqn:HBR}
    R_{HB} = R_{sat}+R_{deb}.
\end{equation}
The sphere with radius $R_{HB}$ is the conjunction space. We denote its volume by $  V_{conj} = \frac{4}{3}\pi R_{HB}^3$.

\textbf{Probability of Collision (PoC)}. We use  Foster's method in~\citep{Foster1992AVehicles} to compute the PoC of short-term LEO encounters. This method is widely used for space missions and is a fundamental component in NASA's Conjunction Assessment Risk Analysis (CARA) program~\citep{2023NASAIi}. Let $\boldsymbol{\Sigma} = \boldsymbol{\Sigma}_{sat}+\boldsymbol{\Sigma}_{deb}$ be the combined position covariance of target satellite and chasing debris. Then, the PoC is defined by the volume integral of the 3D PDF from Eq.~\eqref{eqn:3DPDF} over the conjunction space $V_{conj}$  centered on the chasing debris as
\begin{equation}\label{eqn:3DPOC}
    P_C(\boldsymbol{\rho}_r, \boldsymbol{\Sigma}) = \frac{1}{\sqrt{(2\pi)^3\det(\boldsymbol{\Sigma})}} \int_{V_{conj}} \exp\left[-\frac{1}{2}\boldsymbol{\rho}_r^\top\boldsymbol{\Sigma}^{-1}\boldsymbol{\rho}_r\right]dV,
\end{equation}
In~\citep{Foster1992AVehicles}, the PoC expression is simplified into a 2D integral by projecting the the combined position covariance  $\boldsymbol{\Sigma}$  onto a conjunction plane $\mathbf{\mathcal{B}}$, whose basis vectors are given by
\begin{equation}
\hat{\boldsymbol{x}}_{\bB}=\frac{\boldsymbol{\rho}_r}{\norm{\boldsymbol{\rho}_r}},\hspace{0.5in}\hat{\boldsymbol{y}}_{\bB}=\frac{\boldsymbol{\rho}_r\times\boldsymbol{v}_r}{\norm{\boldsymbol{\rho}_r\times\boldsymbol{v}_r}}.
\end{equation}

A visual representation of the $\bB$-plane is shown in Fig.~\ref{fig:conj_plane} below. 
To project vectors from the RTN-frame to the $\bB$-plane, we use a projection matrix $\boldsymbol{M}_{\bB}  = \begin{bmatrix}
        \hat{\boldsymbol{x}}_{\bB}& \hat{\boldsymbol{y}}_{\bB}
    \end{bmatrix}^\top \in \reals^{2\times3}$.
The $\bB$-plane projection of the combined position covariance  $\boldsymbol{\Sigma}$ is then given by 
\begin{equation}\label{eqn:2Dcov}
    \boldsymbol{\Sigma}_{\bB} = \boldsymbol{M}_{\bB}\boldsymbol{\Sigma} \boldsymbol{M}_{\bB}^\top = \begin{bmatrix}
            \sigma_x^2 & \sigma_{xy} \\ \sigma_{xy} & \sigma_y^2
        \end{bmatrix} \in \reals^{2\times 2}.
\end{equation}
Furthermore, the relative position vector $\boldsymbol{\rho}_r$ expressed in RTN can be projected onto the $\bB$-plane as $\boldsymbol{\rho}_{r,\bB} 
= [\rho_r,0]^\top$ where $\rho_r$ is the miss distance at TCA. The 2D PoC integral is then obtained by combining Eq.~\eqref{eqn:3DPOC} and Eq.~\eqref{eqn:2Dcov} as
\begin{equation}\label{eqn:2DPOC}P_C(\boldsymbol{\rho}_{r,\bB},\boldsymbol{\Sigma}_{\bB}) = \frac{1}{2\pi\sqrt{\det(\boldsymbol{\Sigma}_{\bB})}}\int_{-R_{HB}}^{+R_{HB}} \int_{-\sqrt{R_{HB}^2-x_\bB^2}}^{+\sqrt{R_{HB}^2-x_\bB^2}}\exp\left[-\frac{1}{2}\boldsymbol{\rho}_{r,\bB}^\top \boldsymbol{\Sigma}_{\bB}^{-1}\boldsymbol{\rho}_{r,\bB}\right]dy_\bB dx_\bB.
\end{equation} 
Despite its PoC prediction accuracy, Foster's method is slow to compute, and has step size-dependent precision bounds~\citep{li_review_2022}. We follow ~\citep{Alfriend1999ProbabilityAnalysis} to approximate Eq.~\eqref{eqn:2DPOC} by assuming constant probability density over the collision sphere. The PoC can thereby be approximated as
\begin{equation}\label{eqn:approx_poc}
    P_C (\boldsymbol{\rho}_{r,\bB},\boldsymbol{\Sigma}_{\bB})\approx\frac{R_{HB}^2}{2\sqrt{\det(\boldsymbol{\Sigma}_{\bB})}}\exp\left[ -\frac{1}{2}\boldsymbol{\rho}_{r,\bB}^\top(\boldsymbol{\Sigma}_{\bB})^{-1}\boldsymbol{\rho}_{r,\bB} \right].
\end{equation}

\begin{figure}[h!]
    \begin{minipage}[c]{0.55\linewidth}
    \includegraphics[width=\linewidth]{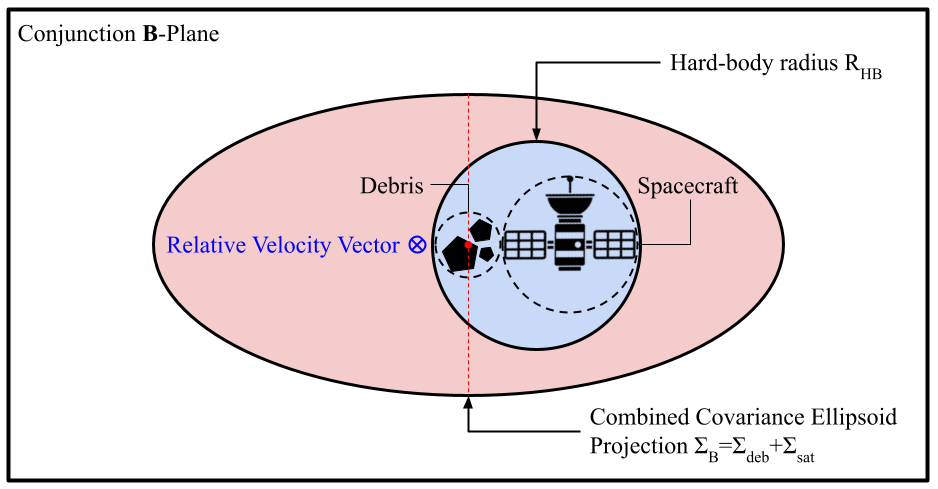}
    \caption{Visualization for a conjunction encounter and conjunction plane illustration.}
    \label{fig:conj_plane}
    \end{minipage}
    \hfill
    \begin{minipage}[c]{0.4\linewidth}
    \includegraphics[width=\linewidth]{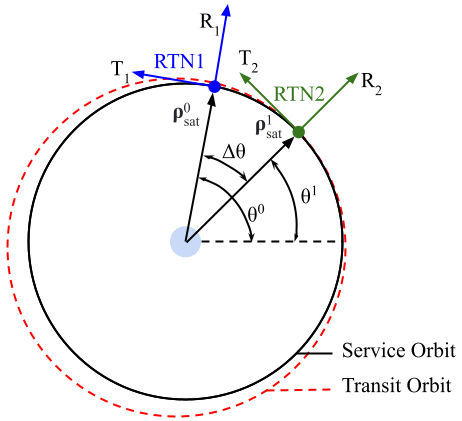}
    \caption{The orbital phase shift and position change at TCA due to the phasing maneuver. Diagram illustrates the phasing maneuver and relates the ECI and RTN coordinate frames.}
    \label{fig:phasing}
    \end{minipage}
\end{figure}

\subsection{CAM---Phasing Maneuvers}\label{sec:CAM} 
Although multiple phasing maneuvers exist and selecting a maneuver is an involved part of CAM operations, we only consider prograde in-track burns in this work since they are preferred for their propellant efficiency and minimal impact on orbital inclination. Additional maneuvers will be considered in future work. 
An in-track burn CAM phasing maneuver requires two impulse-based transfers, the first is a prograde maneuver that places the satellite in a slightly higher transit orbit, and the second maneuver returns the spacecraft to the service orbit. After spending a certain period of time $\Delta t$ in the transit orbit, the orbital periods' differences  $T_{serv} < T_{tran}$ create a phase shift (angular separation) between satellite's position at TCA with and without maneuvering, as shown in Fig.~\ref{fig:phasing}. We denote the satellite's angular positions by $\theta^0, \theta^1$, respectively, centered around the ECI frame as shown in Fig.~\ref{fig:phasing}. This section derives the relationships between PoC, phase shifts, and miss distances. While these  analytical relationships are known ~\citep{Klinkrad2006SpaceAnalysis, Zhang2019DiscreteEnvironment}, we provide proof within our problem setting for completeness. We assume that the CAM trajectory does not introduce additional conjunctions risks. In practice, mission operators finds such a trajectory and is generally feasible~\citep{2023NASAIi}. Both orbits' semi-major axes are approximately their respective radii $R_{serv}$ and $R_{tran}$, defined as the distance between orbit and the center of Earth~\citep{nasa_earthfactsheet}. 


\begin{lemma}[Miss Distance to Phase Shift~\citep{Klinkrad2006SpaceAnalysis, Zhang2019DiscreteEnvironment}]
Let the satellite-debris's relative position at TCA without maneuvering be defined as $\boldsymbol{\rho}_r^{0,RTN1} = [\rho_{r,R}^0,\rho_{r,T}^0,\rho_{r,N}^0]^\top\in\reals^3$ with miss distance $\rho^0_r = \norm{\boldsymbol{\rho}_r^{0,RTN1}}$ on the conjunction plane. In order to increase the miss distance to $\rho_r^1 > \rho_r^0$, the phase shift required is given by
\begin{equation}\label{eqn:final_dtheta}
        \Delta \theta = \frac{1}{R_{serv}}\left(-\rho^0_{r,T}+\sqrt{(\rho^0_{r,T})^2-((\rho_r^0)^2-(\rho_r^1)^2)}\right).
    \end{equation}
\end{lemma}
\begin{proof}
    As shown in Fig.~\ref{fig:phasing}, at TCA and relative to Earth, let $\boldsymbol{\rho}_{sat}^0,\boldsymbol{\rho}_{sat}^1 \in \reals^3$ be the satellite's position without maneuvering in RTN1 frame and with maneuvering in RTN2 frame, respectively.
    By definition, $\boldsymbol{\rho}_{sat}^{0,RTN1} = 
    \boldsymbol{\rho}_{sat}^{1,RTN2} = [R_{serv},0,0]^\top$. 
    Their relative position is $\Delta\boldsymbol{\rho}_{sat} = \boldsymbol{\rho}_{sat}^1 - \boldsymbol{\rho}_{sat}^0$.  
    We rotate $\boldsymbol{\rho}_{sat}^0$ from RTN1 to RTN2 via a direct cosine matrix  $\boldsymbol{R}^{RTN2}_{RTN1}(\Delta\theta)\in\reals^{3\times 3}$~\citep{DumbleRotationMatrices}, where $\Delta\theta$ is the phase shift achieved by the prograde phasing maneuver, $\Delta\theta = \theta^0-\theta^1>0$, so that $\Delta\boldsymbol{\rho}_{sat}$ in RTN2 frame is given by
    \begin{equation}
    \Delta\boldsymbol{\rho}_{sat}^{RTN2} = \boldsymbol{\rho}_{sat}^{1,RTN2} - \boldsymbol{R}^{RTN2}_{RTN1}(\Delta\theta)\boldsymbol{\rho}_{sat}^{0,RTN1}= \boldsymbol{\rho}_{sat}^{1,RTN2} - \begin{bmatrix}
        \cos\Delta\theta&-\sin\Delta\theta&0\\\sin\Delta\theta&\cos\Delta\theta&0\\0&0&1
    \end{bmatrix}\boldsymbol{\rho}_{sat}^{0,RTN1}
        =\begin{bmatrix}
            R_{serv}(1-\cos\Delta\theta)\\-R_{serv}\sin\Delta\theta\\0
        \end{bmatrix}.
    \end{equation}
    Since the phase shift $\Delta \theta$  is small relative to service orbit radius $R_{serv}$, we use the small angle approximation to approximate $\Delta\boldsymbol{\rho}_{sat}^{RTN2}$ as $[0,-R_{serv}\Delta\theta,0]^\top$.
    Next, given the debris-satellite's relative position $\boldsymbol{\rho}_r^{0,RTN1}$ without maneuvering, the relative position $\boldsymbol{\rho}_r^{1,RTN1}$ with maneuvering is given by
    \begin{align}
        \boldsymbol{\rho}_r^{1,RTN1} = &\boldsymbol{\rho}_{deb}^{1,RTN1} - \boldsymbol{\rho}_{sat}^{1,RTN1},\\
        =&\boldsymbol{\rho}_r^{0,RTN1}-\Delta\boldsymbol{\rho}_{sat}^{RTN1}, \label{eqn:pf_1}\\
        =&\boldsymbol{\rho}_r^{0,RTN1} - \boldsymbol{R}^{RTN1}_{RTN2}(\Delta\theta)\Delta\boldsymbol{\rho}_{sat}^{RTN2}, \label{eqn:pf_2}\\
        =&\begin{bmatrix}
            \rho_{r,R}^0&\rho_{r,T}^0&\rho_{r,N}^0
        \end{bmatrix}^\top+\begin{bmatrix}
            R_{serv}\Delta\theta^2&R_{serv}\Delta\theta&0
        \end{bmatrix}^\top,
    \end{align}
    where we added term $-\boldsymbol{\rho}_{sat}^{0,RTN1}+\boldsymbol{\rho}_{sat}^{0,RTN1}$ to derive Eq.\eqref{eqn:pf_1}, and used the frame rotation matrix $\boldsymbol{R}^{RTN1}_{RTN2}(\Delta\theta)$ to derive the RTN1 frame $\Delta\boldsymbol{\rho}_{sat}$ in Eq.\eqref{eqn:pf_2}. 
    By the small angle approximation, $\Delta\theta^2 << \Delta\theta$, 
    such that $\boldsymbol{\rho}_r^{1,RTN1} = \begin{bmatrix}
            \rho_{r,R}^0& \rho_{r,T}^0 + R_{serv}\Delta\theta& \rho_{r,N}^0
        \end{bmatrix}^\top$.
    We  evaluate the miss distance with maneuvering as $\norm{\boldsymbol{\rho}_r^{1,RTN1}} = (\rho_r^1)^2 = (\rho_{r,R}^0)^2+(\rho_{r,T}^0 + R_{serv}\Delta\theta)^2 + (\rho_{r,N}^0)^2$.
    Rearranging this expression produces a quadratic expression  $R_{serv}^2\Delta\theta^2 + 2\rho_{r,T}^0R_{serv}\Delta\theta+((\rho_r^0)^2-(\rho_r^1)^2)=0$. Solving  for $\Delta\theta$ produces $\Delta\theta = \frac{1}{R_{serv}}\Big(-\rho^0_{r,T}\pm\sqrt{(\rho^1_{r,T})^2-((\rho_r^0)^2-(\rho_r^1)^2)}\Big)$.
Since the rotation from RTN1 to RTN2 is clockwise, $\Delta\theta\geq0$~\citep{DumbleRotationMatrices}. Therefore, we take the positive root to derive Eq.~\eqref{eqn:final_dtheta}.
\end{proof}

\begin{lemma}[Threshold PoC to Miss Distance Requirement]
Let $P^i_C = P_C(\rho^i_r, \boldsymbol{\Sigma}_\bB)$ be the PoC at TCA for miss distances $\rho^i_r \in \reals_+$ without $(i=0)$ and with $(i=1)$ maneuvering. If $P^1_C \leq P^0_C/\lambda$ for a scalar safety factor $\lambda \geq 1$, then  miss distance $\rho^1_r$ satisfies
    \begin{equation}\label{eqn:safe_missd}
        (\rho^1_r)^2 \geq\frac{2\det(\boldsymbol{\Sigma}_\bB)}{\sigma_y^2}\ln{\left( \frac{R_{HB}^2\lambda}{2\sqrt{\det(\boldsymbol{\Sigma}_{\bB})}P^0_C}\right)},
    \end{equation}
where $\boldsymbol{\Sigma}_{\bB}$ (Eq.~\eqref{eqn:2Dcov}) is the combined position covariance, $R_{HB}$ is the hard-body radius, and $\sigma_y$ is the $(y,y)$-component of $\boldsymbol{\Sigma}_{\bB}$. 
\end{lemma}
\begin{proof}
Let PoC without maneuvering be denoted as $P_C^0 = P_C(\boldsymbol{\rho}^0_{r,\bB},\boldsymbol{\Sigma}_\bB)$ (Eq.~\eqref{eqn:approx_poc}), where the relative position vector is $\boldsymbol{\rho}^0_{r,\bB} = [\rho_r^0,0]^\top\in\reals^2$ on the conjunction plane.
We  explicitly evaluate $\boldsymbol{\rho}_{r,\bB}^\top\Sigma_\bB^{-1}\boldsymbol{\rho}_{r,\bB} $ from Eq.~\eqref{eqn:approx_poc} as 
    \begin{equation}\label{eqn:poc_num}
    \boldsymbol{\rho}_{r,\bB}^\top\boldsymbol{\Sigma}_\bB^{-1}\boldsymbol{\rho}_{r,\bB} = \begin{bmatrix}
        \rho_r^0 & 0
    \end{bmatrix}\left(\frac{1}{\det(\boldsymbol{\Sigma}_\bB)}\begin{bmatrix}
        \sigma_y^2 & -\sigma_{xy} \\ -\sigma_{xy} & \sigma_x^2
    \end{bmatrix}\right)\begin{bmatrix} \rho_r^0 \\ 0 \end{bmatrix} = \frac{\sigma_y^2(\rho_r^0)^2}{\det{(\boldsymbol{\Sigma}_\bB)}}. 
    \end{equation}
If the miss distance with maneuvering is given by  $\rho_r^1$ at TCA, then the corresponding PoC can be evaluated via Eq.~\eqref{eqn:approx_poc} as
    \begin{equation}\label{eqn:new_poc}
        P^1_C = P_C(\boldsymbol{\rho}^1_{r,\bB},\boldsymbol{\Sigma}_\bB)=\frac{R_{HB}^2}{2\sqrt{\det(\boldsymbol{\Sigma}_\bB)}}\exp\left[-\frac{\sigma_y^2(\rho_r^1)^2}{2\det{(\boldsymbol{\Sigma}_\bB)}} \right].
    \end{equation}
Let $P^1_C \geq P^0_C/\lambda$ and combine it with Eq.~\eqref{eqn:new_poc}. We can solve for the miss distance $\rho_r^1$ to derive Eq.~\eqref{eqn:safe_missd}.
\end{proof}

\subsection{Propellant Consumption under High Thrust Propulsion}\label{sec:FUEL} 
In this section, we determine the propellant consumption required to guarantee a safe miss distance at TCA with maneuvering.
In a prograde phasing maneuver, the satellite will carry out $n_r$ revolutions in 
a higher circular orbit that we denote as the transit orbit.
The satellite accumulates phase shift by passively revolving in the transit orbit over time. Therefore, increasing the revolution time can decrease the transit orbit radius while achieving the same phase shift $\Delta\theta$.
Let the transit orbit radius be denoted as $R_{tran}$ and the satellite's speed in transit orbit be denoted as $V_{tran}$~\citep{Kluever2018SpaceDynamics}. These quantities are related via orbital mechanics~\citep{nasa_earthfactsheet} as
\begin{equation}
    \textstyle R_{tran}=\left(\frac{\mu T_{tran}^2}{4\pi^2}\right)^{1/3},\quad V_{tran}=\sqrt{\frac{\mu}{R_{tran}}}, \quad \mu=0.3986\times10^6 km^3/s^2.
\end{equation}
\begin{lemma}[Transit Orbit Period as a Function $\Delta\hat{\theta}$ and $n_r$~\citep{WeberOrbitalMechanics}]\label{lem:trans_orbit}
If the transit orbit of a prograde phasing maneuver produces $\Delta\theta \in\reals_+$ phase shift in $n_r \in \mathbb{Q}_+$ revolutions, then the transit orbit's period $T_{tran}$ must satisfy  
\begin{equation}\label{eqn:transit_period}
    T_{tran}\geq T_{serv}+\frac{R_{serv}\Delta\theta}{n_rV_{serv}},
\end{equation}
where $R_{serv}$, $V_{serv}$, and $T_{serv}$ denote the radius, speed, and period of the satellite's service orbit respectively.
\end{lemma}\begin{proof}
Let $T_i\in\reals_+,\,i\in\{serv,tran\}$ denote the service and transit orbits' periods respectively, 
and let $\dot{\theta}_{i} = V_i/R_i,\,i=\{serv,tran\}$ denote the service and transit orbits' constant angular speeds, respectively. If within the total phasing time $\Delta t $, a phase shift of $\Delta\theta^1$ is achieved, then  $\Delta t = n_rT_{tran} = \Delta\theta^1/\dot{\theta}_{serv}+n_rT_{serv}.$
We substitute in $\dot{\theta}_{serv} = V_{serv}/R_{serv}$  to derive $n_rT_{tran}=\frac{R_{serv}\Delta\theta^1}{V_{serv}}+n_rT_{serv}$. Solve for $T_{tran}$ produces $T_{tran}=\frac{R_{serv}\Delta\theta^1}{V_{serv}n_r}+T_{serv}$. If $\Delta\theta^1 \geq \Delta\theta$, we arrive at the inequality Eq.~\eqref{eqn:transit_period}. 
\end{proof}
We model the CAM as an impulsive maneuver, where 
the satellite's speed changes instantaneously in between orbits.
The total speed change for the overall CAM to and from the transit orbit is $\Delta V = 2(V_{serv} - V_{tran})$. 
To convert the speed change to propellant mass, we utilize  a high-thrust propulsion model. Even though LEO satellites typically have low-thrust propulsion systems onboard, we make this simplifying assumption here, and plan to explore the low-thrust alternative in future works. We model the propellant consumed $m_p$ by the high-thrust propulsion systems via the 
Tsiolkovsky equation~\citep{Walter2018Astronautics} rearranged as
\begin{equation}\label{eqn:fuel}
    m_p = m_0\left[ 1-\exp{\left(\frac{-\Delta V}{I_{sp}\,g_0}\right)} \right],
\end{equation}
where $m_0$ is the satellite's initial (pre-maneuver) mass, $I_{sp}$ is the satellite's propulsion system's specific impulse, and $g_0$ is the gravitational acceleration constant at sea-level.

\section{Conjunction Data Dynamics via MDP}\label{sec:mdp}
This section casts the satellite CAM problem into a finite horizon MDP. 
MDPs are stochastic decision-making models in which a taken action influences the immediate reward, future states, and thus future rewards~\citep{Puterman2008MarkovProgramming}. These frameworks are essential in situations such as collision avoidance, where the maneuvering decision is made over a period of time under uncertainty. In this work we implement a continuous state, discrete action, finite-horizon MDP. The policy represents a mapping from dynamical states to discrete actions (i.e. maneuver or stay). To learn the trained policy, we use an RL-PG algorithm called \textit{REINFORCE} that directly computes an trained policy based on observed trajectories from environment interactions~\citep{Williams1992SimpleLearning}.


We utilize a finite horizon, continuous state, and discrete action MDP defined by the tuple $(\mathcal{S}, \mathcal{A}, \mathcal{P}, \ell, K)$. Here, $K = 21$ is the MDP time horizon, $\mathcal{S} \subset \mathbb{R}^2$ is the state space, $\mc{A}\in \mathbb{N}_+$ is the action space, $\mc{P}:\mc{S}\times \mc{A} \times [K] \mapsto \Delta_{\mc{S}}$ defines the time-dependent transition probabilities where $s_{k+1} \sim \mathcal{P}(s_k, a_k, k)$, and $\ell: \mc{S}\times\mc{A}\times [K] \mapsto \reals$ is the time-state-action-dependent reward. Within this MDP framework, we model the conjunction data as MDP states with memory-less evolution of its values 
via stochastic Markov dynamics~\citep{caldas2023conjunction}. We assume that the updates to observed conjunction parameters defined in the CDMs (e.g. position covariances, velocity) are stochastic and follow time-dependent probability distributions. While this is supported by~\citep{caldas2023conjunction}, accurately predicting collision risk is challenging and an active research area in and of itself \citep{KelvinsHome,Balch2016AAnalysis}.

\textbf{State Space.}\label{sec:mdp_states}
Ideally, all CDM attributes would be included. However, a CDM can contain up to $90$ relevant attributes and including all in the state space can unnecessarily complicate the dynamics model~\cite{2023NASAIi}. To balance model fidelity with computation efficiency, we focus on the following subset of CDM attributes in the MDP state space:
\begin{enumerate}
    \item $\rho \in [0, 100km]$: the miss distance between the satellite and the debris;
    \item $\sigma \in [0, 100km]$: the debris' along-track standard deviation, equivalent to the square root of the tangential component of $\boldsymbol{\Sigma}_{deb}$ expressed in the RTN-frame.

\end{enumerate}
These two variables are the most immediately relevant to the PoC and CAM computations, and the most correlated to PoC uncertainty.
We chose to highlight the along-track (tangential) standard deviation $\sigma$ of the debris because it
tends to be several orders of magnitude larger than the debris and satellite covariance terms in the radial and normal directions.
We model each relevant CDM attribute as a one dimensional stochastic variable with independent transition dynamics~\citep{caldas2023conjunction}. To model the state of a post-maneuver satellite, we introduce a binary state variable, $moved$, indicating whether a CAM has been initiated. The resulting state space is
\begin{equation}
\begin{bmatrix}
    \rho &  \sigma & moved 
\end{bmatrix} \in \mc{S} = [0, 100km]^{2} \times \{0,1\}.
\end{equation}

\textbf{Action Space}. When the action $moved = True$, the satellite has initiated the CAM and must continue. When the action $moved = False$, the satellite can decide whether to delay further or maneuver immediately. At each time step $k$ and only in states $s_k$ where $moved=False$, the available actions are to maneuver this time step, or delay the decision until the next time step:
\begin{equation}
    \mathcal{A}= \: \{ delay, maneuver\} = \{0,1\}.
\end{equation}
Selecting the \textit{delay} action ($a=0$) lets the satellite continue orbiting without intervention, consuming no propellant. Selecting the \textit{maneuver} action ($a=1$) indicates a CAM initialization at the given time step.

\textbf{Transition Dynamics}\label{sec:MDP_transitions}
We assume that all states evolve independently over time. The binary state $moved$ remains $0$ until a $maneuver$ action is taken, then remains $1$ throughout the time horizon. We assume that the updates to the CDM conjunction parameters are stochastic and follow time-dependent probability distributions.
While this is supported by~\citep{caldas2023conjunction}, accurately predicting collision risk is challenging and an active research area in and of itself \citep{KelvinsHome,Balch2016AAnalysis}.

After exploring both additive and multiplicative noise models, we determined that multiplicative noise best models the stochastic revolution of both miss distance and the tangential covariance of the chaser satellite. Miss distance is best represented by the dynamics $\rho_{k+1} = \rho_k (1 + w_k^\rho ),$  where $w_k^\rho$ is a generalized normal distribution (GND)~\citep{nadarajah2005generalized}, defined as $w_k^\rho \sim \text{GND}(\mu_k^\rho, \alpha_k^\rho, \beta_k^\rho)$, with time-dependent location $\mu_k^\rho$, scale $\alpha_k^\rho > 0$ and shape $\beta_k^\rho > 0$ parameters. Let $\Gamma$ represent the gamma function. Then, the GND's pdf is given by
\begin{equation}\label{eqn:gnd}
f(x|\mu, \alpha, \beta) = \frac{\beta}{2 \alpha \Gamma(1/\beta)} \exp\left(-\left(\frac{|x - \mu|}{\alpha}\right)^\beta\right).
\end{equation}

We also found that the along-track standard deviation $\sigma$, is best represented by the dynamics $\sigma_{k+1} = \sigma_k (1 + w_k^\sigma)$,
where $w_k^\sigma$ empirically follows a non-central t-distribution (NCT)~\citep{kay1993fundamentals} i.e. $ w_k^\sigma \sim \text{NCT}(\nu_k^\sigma, \delta_k^\sigma)$, with time-dependent number of degrees of freedom $\nu_k^\sigma > 0$ and non-centrality parameter $\delta_k^\sigma  \in \mathbb{R}$.
The NCT is a generalization of the Student's t-distribution, and its pdf is given as
\begin{equation}\label{eqn:nct}
f(k|\nu, \delta) = \frac{\nu^{\nu/2} \Gamma\left(\frac{\nu+1}{2}\right)}{\sqrt{\pi \nu} \Gamma\left(\frac{\nu}{2}\right) (1 + k^2/\nu)^{(\nu+1)/2}} \cdot e^{-\delta^2/2} \sum_{r=0}^{\infty} \frac{\Gamma\left(\frac{\nu+r+1}{2}\right)}{\Gamma(r+1)} \left(\frac{\delta k}{\sqrt{\nu (1+k^2/\nu)}}\right)^r.
\end{equation}

\textbf{Time-dependent Rewards}\label{sec:rewardsmdp}. 
The reward function is the main driver of the learning process towards desired behaviors~\citep{Sutton2018ReinforcementIntroduction}. It balances two objectives: propellant consumption $m_p$ (Eq.~\eqref{eqn:fuel}) and collision risk $P_C$ (Eq.~\eqref{eqn:approx_poc}). 
As shown in Eq.~\eqref{eqn:fuel}, the propellant consumption $m_p$ depends on the required instantaneous speed change $\Delta V$. Using Lemma~\ref{lem:trans_orbit}, we can show that $m_p$ subsequently depend on the required phase shift $\Delta \theta$, the number of transit orbit revolutions $n_r$, and the transit orbit chosen $T_{tran}$. The required phase shift $\Delta \theta$ is computed using MDP states via Eq.~\eqref{eqn:final_dtheta} and Eq.~\eqref{eqn:safe_missd},  the number of revolutions $n_r$ decreases with increasing MDP time step $k$, and the transit orbit $T_{tran}$ is selected to satisfy Eq.~\eqref{eqn:transit_period}.
Let $\bar{m}_p(s, k) \in [0.0,1.0]$ be the propellant consumption at time-step $k$ normalized with respect to the worst case propellant consumption. Then, we compute the time-dependent propellant reward as \begin{equation}\label{eqn:prop_reward}
    \ell_{prop}(s,a,k) = \begin{cases}
        0.0 & a=0,\\ -\bar{m}_p(s,k) & a=1.
    \end{cases}, \quad \forall (s,a,k ) \in \mc{S}\times\mc{A}\times [K-1].
\end{equation}
We compute the PoC at final timestep $K$ using state information via Eq.~\eqref{eqn:approx_poc}, i.e. $P_C: \mc{S} \mapsto \reals_+$. For a safe PoC threshold $P_{C,max} \in \reals_+$, the reward of collision risk is defined by 

\begin{equation}\label{eqn:risk_reward2}
    \ell_{risk}(s) = \begin{cases}
        -10.0 & moved=False,\,P_C(s)\geq P_{C,max},\\
        -5.0 & moved=True,\,P_C(s)< P_{C,max},\\
        0.5 & moved=False,\,P_C(s)< P_{C,max},\\
        1.0 & moved=True,\,P_C(s)\geq P_{C,max}.
    \end{cases}, \quad \forall s \in \mc{S}.
\end{equation}
Because the miss distance and along track standard deviation propagate via transition dynamics~\eqref{eqn:gnd} and~\eqref{eqn:nct} regardless of the satellite's maneuver decision, these states $(\rho_k, \sigma_k)$ always model the collision risk if the satellite \emph{did not maneuver}.
For a realized state-action trajectory $(s_0, a_0, \ldots, s_K)$, the finite horizon MDP reward trades off between the propellant consumption and the collision risk  via a weighting factor $\eta \in [0,1]$, and is given by 
\begin{equation}\label{eqn:reward}
    \sum_{k=0}^{K} \ell(s_k, a_k, k) = (1-\eta) \ell_{risk}(s_K) + \eta \sum_{k=0}^{K-1}\ell_{prop}(s_k, a_k, k).
\end{equation}

\textbf{Markov Policy}. 
We consider policies that are strictly functions of the current state $s_k$ and time step $k$~\citep{Sutton2018ReinforcementIntroduction}. 
We denote the policy by $\pi$, given by 
\begin{equation}\label{eqn:policy}
\pi(a|s, k) = \mathbb{P}(a_k = a | s_k = s, k), \quad \forall (s,a,k)\in \mc{S} \times \mc{A}\times[K-1]. 
\end{equation}
Each policy is a conditional probability distribution over the action space $\mc{A}$, so that $\pi(a|s,k)$ is the probability of selecting action $a\in\mathcal{A}$ at state $s$ and time step $k$.
Finally, we parametrize the policy $\pi$ by $\boldsymbol{\phi} \in \reals^{m}$ to facilitate training policy $\pi_{\boldsymbol{\phi}}$ over parameters $\boldsymbol{\phi}$.
The trained policy minimizes the expected propellant and collision risk rate-balanced reward of a conjunction event given by
\begin{equation}\label{eqn:objective}
   J (\pi_{\boldsymbol{\phi}}) = \mathbb{E} \left[ \sum_{k=0}^{K} \ell(s_k, a_k, k) | a_k \sim \pi_{\boldsymbol{\phi}}(s_k, k) \right].
\end{equation}

Given historical CDM data, we perform gradient descent on the policy parameters $\boldsymbol{\phi}$ to minimize the $\log$ transformation of Eq.~\eqref{eqn:objective}, whose gradient is given by
\begin{equation}\label{eqn:nablaJ}
       \nabla_{\boldsymbol{\phi}} J(\pi_{\boldsymbol{\phi}}) = \mathbb{E}_{\pi} \left[ \sum_{k=0}^K \nabla_{\boldsymbol{\phi}} \log \pi_{\boldsymbol{\phi}}(a_k|s_k, k)\sum_{k=0}^{K} \ell(s_k, a_k, k) | a_k \sim \pi_{\boldsymbol{\phi}}(s_k, k)\right].
\end{equation} 
We then employ the ``REINFORCE" RL-PG algorithm~\citep{Sutton2018ReinforcementIntroduction} to perform gradient descent  over the policy parameters $\boldsymbol{\phi}$ and optimize the expected finite horizon objective $J (\pi_{\boldsymbol{\phi}})$.  
\begin{figure}[h!]
    \centering
    \begin{minipage}{0.45\textwidth}
        \begin{algorithm}[H]
        \caption{RL-PG ``REINFORCE" Algorithm with Epsilon-Greedy Exploration.}\label{algo:reinforce}
        \begin{algorithmic}[1]
            \Require Initial policy parameters $\boldsymbol{\phi}_0$, exploration rate $\epsilon\in[\epsilon_{\min},\epsilon_{\max}]> 0$, decay rate $\lambda > 0$, and learning rate $\alpha > 0$
            \For{iteration $i = 1$ to $N_{\text{iterations}}$}
            \State Initialize: \textit{states, times, risks, actions, rewards}
                \For{episode $e=1$ to $N_{\text{episodes}}$}
                    \State Reset environment: $s_0 \sim \mathcal{S}$, $\ell_i \gets 0$, $k \gets 0$
                    \While{$k < K$}
                        \State Sample $r \sim \text{Uniform}(0,1)$
                        \If{$r < \epsilon_i$}
                            \State $a_k \sim \text{Uniform}(\mathcal{A})$
                        \Else
                            \State $a_k \sim \pi_\phi(s_k, k)$
                        \EndIf
                        \State Execute $(s_k, a_k)$, observe $s_{k+1}$, 
                        \State $\ell_i \gets \ell_i + \eta \cdot \ell_{prop}(s_k, a_k, k)$ (Eq.~\eqref{eqn:prop_reward})
                        \If{$k = K$}
                            \State $\ell_i \gets \ell_i + (1 - \eta) \cdot \ell_{risk}(s_K, K)$ ( Eq.~\eqref{eqn:risk_reward2})
                            
                        \EndIf
                        \State $s_k \gets s_{k+1}$
                        \State $k \gets k + 1$

                    \EndWhile
                    
                \EndFor
                \State Gradient step: $\phi \gets \phi + \alpha \cdot \nabla_\phi \log \pi_\phi(a_k\mid s_k, k) \cdot \ell_i$
                \State Decay exploration rate: $\epsilon_k \gets \max(\epsilon_{\min}, \epsilon_{\max} \cdot \lambda^i)$
            \EndFor
        \end{algorithmic}
        \end{algorithm}
        \end{minipage} 
    \hfill
    \begin{minipage}{0.53\textwidth}
        \flushleft
        \begin{table}[H]
        \begin{tabular}{llll}
            \toprule
            \textbf{Param.} & \textbf{Default} & \textbf{Var. 1 ($R_{HB}/\Delta\theta$)} & \textbf{Var. 2 ($\eta$)} \\
            \midrule
            $\Delta\theta$ & 0.01 & 0.01 or Eq.~\eqref{eqn:final_dtheta} & 0.01 \\
            $R_{HB}$ & 10.0 m & 10.0 m or rand. CDM & 10.0 m \\
            $n_r$ & $21-k$ & $21-k$ & $21-k$ \\
            $\eta$ & 0.25 & 0.25 & $\eta\in[0,1]$ \\
            $R_s$ & 160--2000 km & 160--2000 km & 160--2000 km \\
            $I_{sp}$ & 300.0 s & 300.0 s & 300.0 s \\
            $m_o$ & 300.0 kg & 300.0 kg & 300.0 kg \\
            \bottomrule
        \end{tabular}
        \caption{Parameter values for the default setting and variations.}
        \label{tab:case_params}
        \end{table}
        \begin{minipage}{0.75\textwidth}
            \flushleft
            \begin{table}[H]
            \begin{tabular}{cccc}
                \toprule
                \textbf{NN Size} & \textbf{Exploration } & \textbf{Decay}& \textbf{Learning}
                \\
                & \textbf{Rate} $\epsilon$ & \textbf{Rate} $\lambda$ & \textbf{Rate} $\alpha$\\
                \midrule
                $64\times 128$ & $[0.01, 0.1]$ & $0.999$ & $10^{-4}$ \\
                \bottomrule
            \end{tabular}
            \caption{RL-PG algorithm and neural network hyper-parameters.}
            \label{tab:nn_hypar}
            \end{table}
            \begin{minipage}{0.9333\textwidth}
                \flushleft
                \begin{table}[H]
                \begin{tabular}{lc}
                    \toprule
                    \textbf{ESA Dataset} & \textbf{Count} \\
                    \midrule
                    Total CDMs & 161124 \\
                    Unique Conj. Events & 11155 \\
                    Avg. CDMs per Event & 14 \\
                    \bottomrule
                \end{tabular}
                \caption{ESA CDM dataset.}
                \label{tab:cac_dataset}
                \end{table}
            \end{minipage} 
        \end{minipage}
    \end{minipage}
\end{figure}

\section{parameter-sensitivity Studies: Training sensitivity and policy performance}\label{sec:prelim_res}
In this section, we validate the MDP model and the RL-PG training using publicly available CDM data from the 2019 Kelvins collision avoidance challenge hosted by the European Space Agency (ESA) in 2019~\citep{KelvinsData}. 
We train a maneuver guidance policy via RL-PG, and analyze the algorithm training complexity (Sec.~\ref{sec:training}) and the output policy performance (Sec.~\ref{sec:empir_res}) under variations in MDP parameters and RL-PG hyper-parameters.
\forjournal{
These parameter-sensitivity studies can be grouped first into an in-distribution and out-of distribution studies, each utilizing 100\% and 75\% of the provided CDM data respectively. The parameters used in both case studies were the same fixed values presented in Tab.~\ref{tab:case_params}.}
All case studies' modeling parameters are shown in Tab.~\ref{tab:case_params}; they are representative of the LEO orbital dynamics as well as the average CDM data. 
To evaluate performance, we compare  against a cut-off policy. 
\subsection{Synthetic CDM Generator}
In this section, we construct a synthetic CDM simulator to support policy training. 
There are two main challenges with directly training on the ESA data set: (i) with $15,321$ conjunction events, the ESM dataset does not contain enough data to train RL-PG to convergence, (ii) only $3\%$ of the conjunction events are considered high risk, with the majority of CDMs having $P_C\approx10^{-30}$. Instead, we construct a synthetic CDM generator by fitting the probabilistic models to the miss distance and tangential standard deviation time-varying transition dynamics, and use these models to generate synthetic CDMs. The probabilistic transition models are described in Sec.~\ref{sec:mdp} via Eq.~\eqref{eqn:gnd} and Eq.~\eqref{eqn:nct},  and are fitted for each time step $k \in [K-1]$.  
While higher fidelity CDM simulators already exist~\citep{Acciarini2020SpacecraftProgramming}, we used a custom simulator to perform consistent training and evaluation within our MDP framework. 

\subsection{RL-PG Training Complexity}\label{sec:training}

In this section we explore the RL-PG algorithm complexity by analyzing the convergence rate of the average MDP reward under variations in modeling parameters, as shown in Tab.~\ref{tab:case_params}. 
The maneuver guidance policy is parametrized by a neural network (NN) with two hidden layers of $64$ and $128$ nodes. We summarize in Tab.~\ref{tab:nn_hypar} the hyper-parameters used for the NN, including the $\epsilon$-greedy exploration strategy exploration rate $\epsilon$ and decay rate $\lambda$, as well as the learning rate $\alpha$ for the Adaptive Moment Estimation (ADAM) optimizer~\citep{ADAMoptimizer} used to update the NN weights, following the architecture described in Alg.~\ref{algo:reinforce}. 
By default, all parameter-sensitivity studies train up to $N_{iterations}=4000$ iterations with $N_{episodes} = 200$ episodes per iteration. However, we vary $N_{iterations}$ as needed to ensure training convergence. 
During each iteration $i \in [N_{iterations}]$, the rewards of $N_{episodes}=200$ episodes are collected, averaged, and stored for the current iteration $i$. Then, during each iteration $i$, we evaluate policy convergence via an empirical moving average that approximates the expected MDP reward Eq.~\eqref{eqn:objective}. Given a set of trajectories $ \{(s_k^{i,e}, a_k^{i,e}, \ell_k^{i,e} )\}_{(i \in [N_{iterations}],\,e\in[N_{episodes}],\, k \in [K-1])}$ generated via policy $\pi_{\boldsymbol{\phi}}$ the empirical moving average is given by
\begin{equation}\label{eqn:empirical_reward}
 \hat{J} (\pi^i_{\boldsymbol{\phi}}) = \frac{1}{H}\left[\sum_{i}^{i+H}\sum_{e=1}^{N_{episodes}}\sum_{k=0}^{K} \ell(s^{i,e}_k, a^{i,e}_k, k)
\right], \,\forall i \in [N_{iterations}], \quad H=50.
\end{equation}

\begin{figure}[h!]
    \centering
        \includegraphics[width=0.51\textwidth]{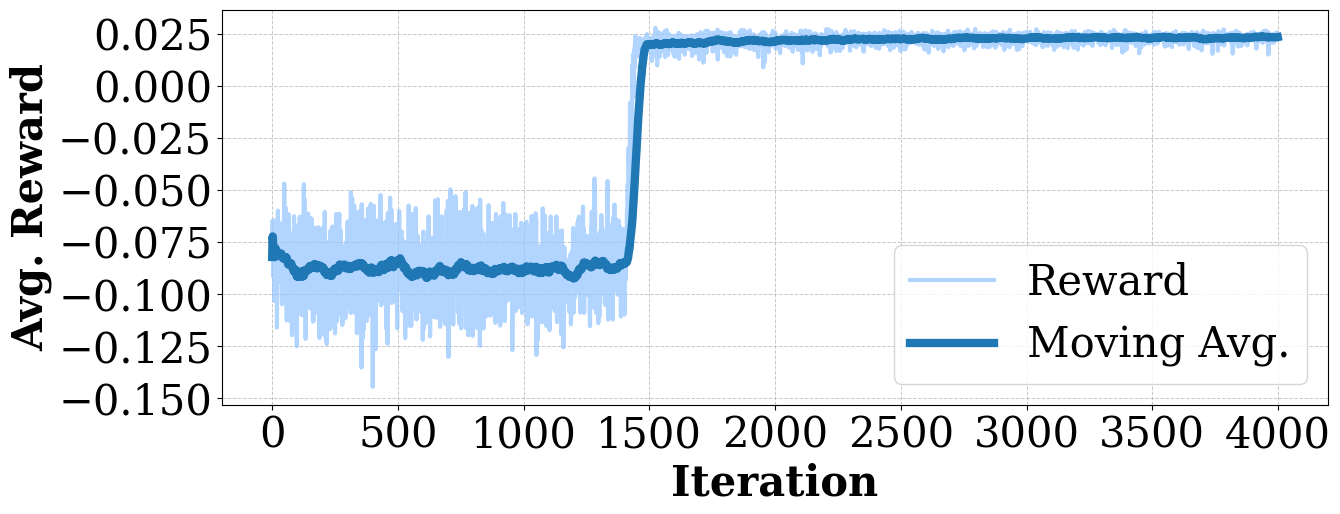}
        \caption{The average reward $\hat{J}(\pi_{\boldsymbol{\phi}}^i)$ (Eq.~\eqref{eqn:empirical_reward})  moving trajectory during Alg.~\ref{algo:reinforce}.}
        \label{fig:convergence}
\end{figure}

\textbf{Default Settings: Convergence Rate}. In this section, we verify the convergence of Alg.~\ref{algo:reinforce} under hyper-parameters defined in Tab.~\ref{tab:nn_hypar} by observing the empirical reward (Eq.~\eqref{eqn:empirical_reward}) convergence  for a policy trained with \textbf{Default} parameters from Tab.~\ref{tab:case_params}. Alg.~\ref{algo:reinforce} converges just before $1500$ iterations in Fig.~\ref{fig:convergence}, where we observe a jump in average reward from approximately $-0.050$ to $0.025$. This sets the baseline for the parameter-sensitivity studies below.

\begin{figure}[h!]
    \centering
    \begin{minipage}{0.35\textwidth}
    \centering
    \begin{table}[H]
        \centering
    
        \begin{tabular}{c c c c}
            \toprule
            \textbf{Index} & \textbf{Fixed}  & \textbf{Fixed}  & \textbf{Convergence}  \\
            $j$ & 
            \textbf{HBR} & \textbf{Phase} & \textbf{Iteration}\\
            \midrule
            1& False & False & \textit{DNC}\\
            2& False & True & $\approx6000$\\ 
            3& True & False & $\approx9200$\\ 
            4& True & True & $\approx1700$\\ 
            \bottomrule
        \end{tabular}
        \caption{All possible combinations of fixed/random HBR and fixed/analytical $\Delta\theta$.}
        \label{tab:binary_options}
        \end{table}
    \end{minipage}%
    \hfill
    \begin{minipage}{0.55\textwidth}
        \includegraphics[width=\textwidth]{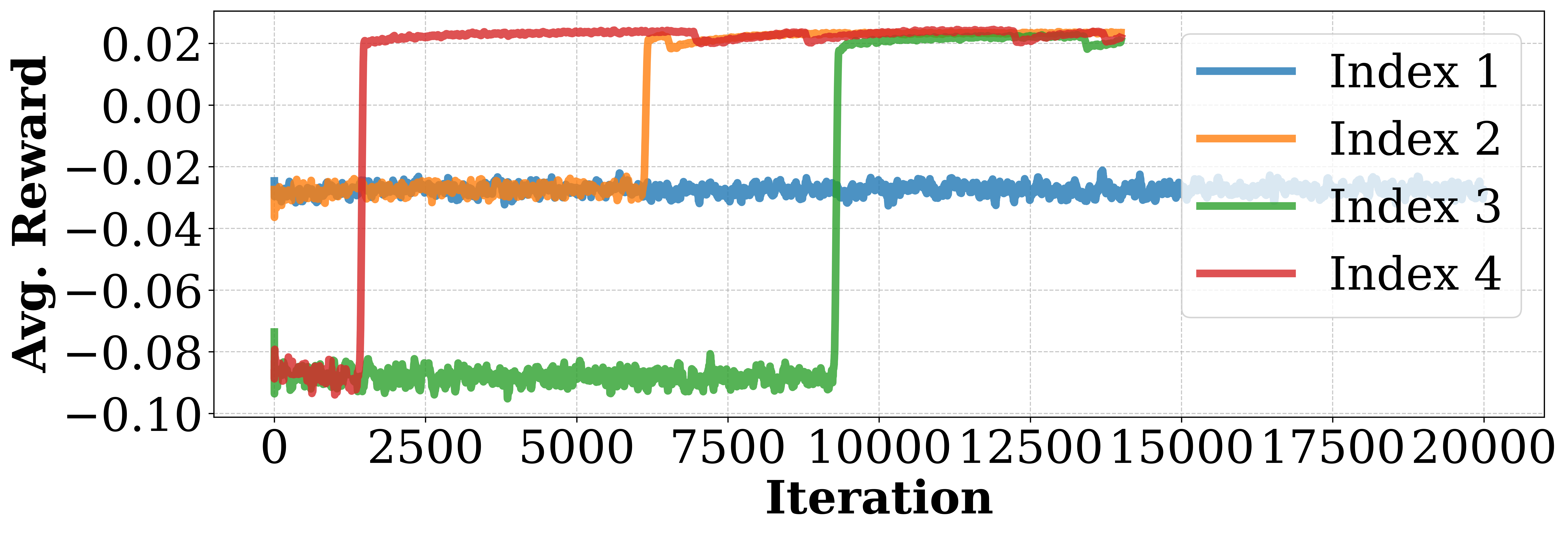}
        \caption{Model verification represented by training convergence via average reward $\hat{J}(\pi^j_{\boldsymbol{\phi}})
        $ (Eq.~\eqref{eqn:empirical_reward}) for each combination of hard-body and phase variations.}
        \label{fig:convergence_binary}
    \end{minipage}
\end{figure}

\textbf{Variation 1: Convergence Rate under Hard Body and Phase Variations}. 
In the default setting, the HBR $R_{HB}$ and phase shift required $\Delta\theta$ are both fixed. However, the $R_{HB}$ can vary depending on the dimensions of the satellite and the debris, and $\Delta\theta$ can vary depending on the conjunction geometry and time horizon. Therefore, in this study we aim to explore how varying $R_{HB}$ and $\Delta\theta$ influences the RL-PG training complexity to determine the combination with the best convergence performance with all other parameters kept as defined in Tab.~\ref{tab:binary_options} under \textbf{Var. 1 $(R_{HB},\Delta\theta)$}. 
Specifically, we explore how all four combinations of $R_{HB}$ and $\Delta\theta$ variations affect the algorithm convergence via average reward $\hat{J}(\pi^j_{\boldsymbol{\phi}}),\,\forall j\in\{1,2,3,4\}$, where we use $j$ to index the modeling parameter set from Tab.~\ref{tab:binary_options} used to train $pi^j_{\boldsymbol{\phi}}$. To make $R_{HB}$ variable, we select $R_{HB}$ from a uniform distribution. 
To make the phase shift $\Delta\theta$ variable, we use Eq.~\eqref{eqn:final_dtheta} to evaluate the required phase shift per CDM.  
Fig.~\ref{fig:convergence_binary} presents the moving average rewards for all four combinations. The fastest convergence occurs under fixed $R_{HB}$ and and fixed $\Delta\theta$. When $R_{HB}$ and $\Delta\theta$ are both variable ($j=1$), Alg.~\ref{algo:reinforce} \textit{does not converge (DNC)}.

\textbf{Variation 2: Convergence Rate under Propellant-risk Trade-off Variations}. 
In this study, we evaluate how different cost weight factor values $\eta \in \reals$ affect the average reward convergence rate.
The MDP reward structure in Eq.~\eqref{eqn:reward} linearly trades off the reward of propellant consumption $\ell_{prop}$ with the reward of collision risk $\ell_{risk}$ via the weighting factor $\eta$. 
Under \textbf{Var. 2 ($\eta$)} parameter definitions from Tab.~\ref{tab:case_params}, we run Alg.~\ref{algo:reinforce} to convergence for eleven $\eta$ values selected from $0$ to $1$ in $0.1$ intervals, and compare their convergence rates. 
We observe that the training convergence rate is positively correlated with increasing values of $\eta\in[0,1]$. 
For $0\leq\eta\leq0.6$, the algorithm convergence rates vary between $[1200,2000]$,  and remain  relatively stable to $\eta$ variations. For $\eta>0.6$, the algorithm convergence rates increase exponentially. When $\eta=1.0$, the algorithm does not converge (DNC) within $20,000$ iterations. We observe that these results make sense intuitively: as $\eta$ increases, more importance is placed on the propellant mass and less on the risk. When $\eta =1.0$, the PoC risk does not affect the MDP reward. Therefore, the most cost-effective policy is to simply not move at all. 

\begin{figure}[ht!]
    \centering
    \includegraphics[width=\linewidth]{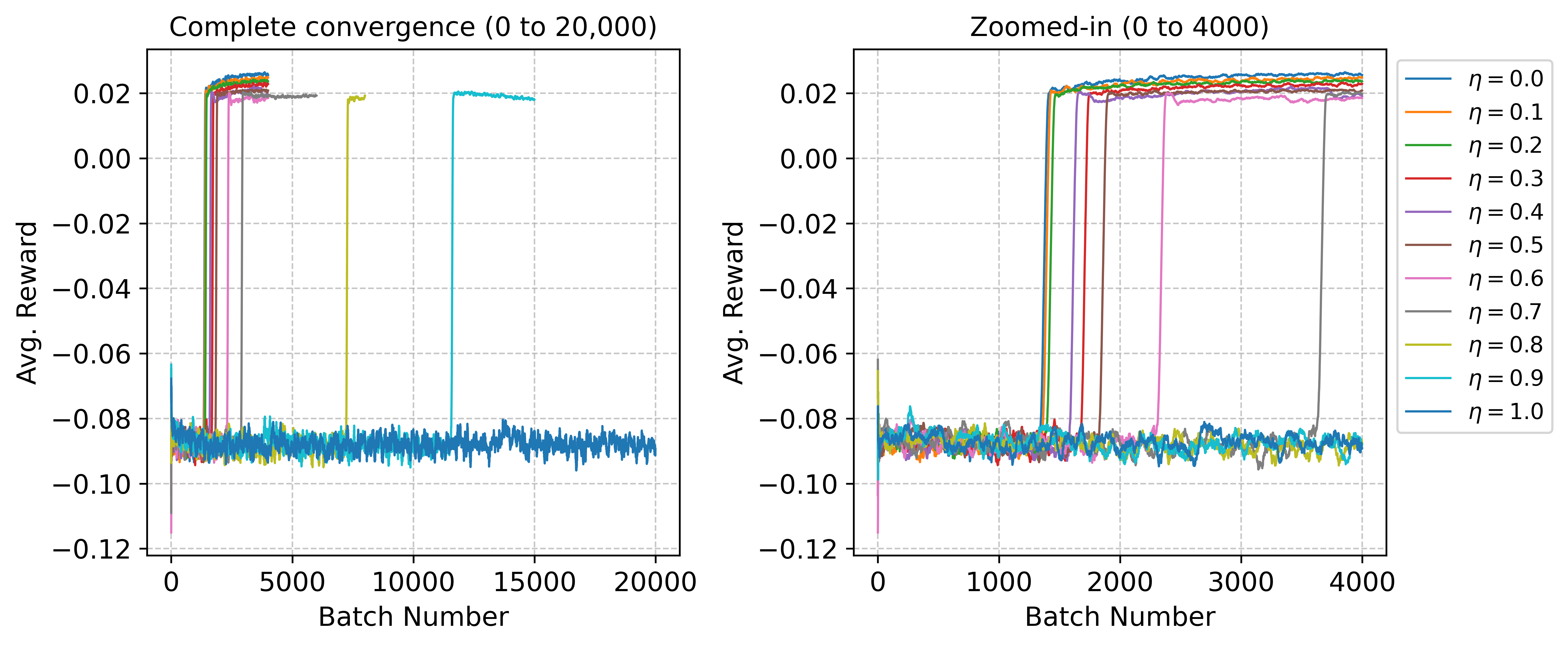}
    \caption{The convergence rate of average reward $\hat{J}(\pi_{\boldsymbol{\phi}}^n)$ (Eq.~\eqref{eqn:empirical_reward}) under varying $\eta$ (Eq.~\eqref{eqn:reward}).}
    \label{fig:eta_conv}
\end{figure}
\forjournal{
\textbf{Average MDP Reward under Optimal Number of Revolutions}. 
In the mathematical models describing the propellant mass minimization, the optimal number of revolutions is defined as $n_r^\star = t/T_t(\min\{R_s+\Delta  R,2000km\})$ where $t$ represents the time remaining to TCA, which defined as $t=168-8k$ for $k\in[0,t_{tot}/8]$ represents the current time step. The remaining parameters involved are presented under \textit{CASE 5} in Tab.~\ref{tab:case_params}. For this case, random HBR and dynamic $\Delta\theta$ are both implemented, which according to Tab.~\ref{tab:binary_options} resulted in non-convergence (within the 8000 iteration limit) for $n_r\leq21$. However, this setup was selected to clearly observe if the optimal $n_r^\star$ parameter would positively impact the performance at all. If the non-convergence case were to converge now, that would signify a clear improvement in the policy training. The resultant average MDP reward as shown in Fig.~\ref{fig:conv_plot_opt} presents a convergence at approximately 6000 iterations, significantly less than the $>8000$ non-convergence of the limited $n_r$ case of Tab.~\ref{tab:binary_options}. This shows that implementing $n_r^\star$ had a positive impact on the policy training.

\begin{figure}
    \centering
    \includegraphics[width=0.5\linewidth]{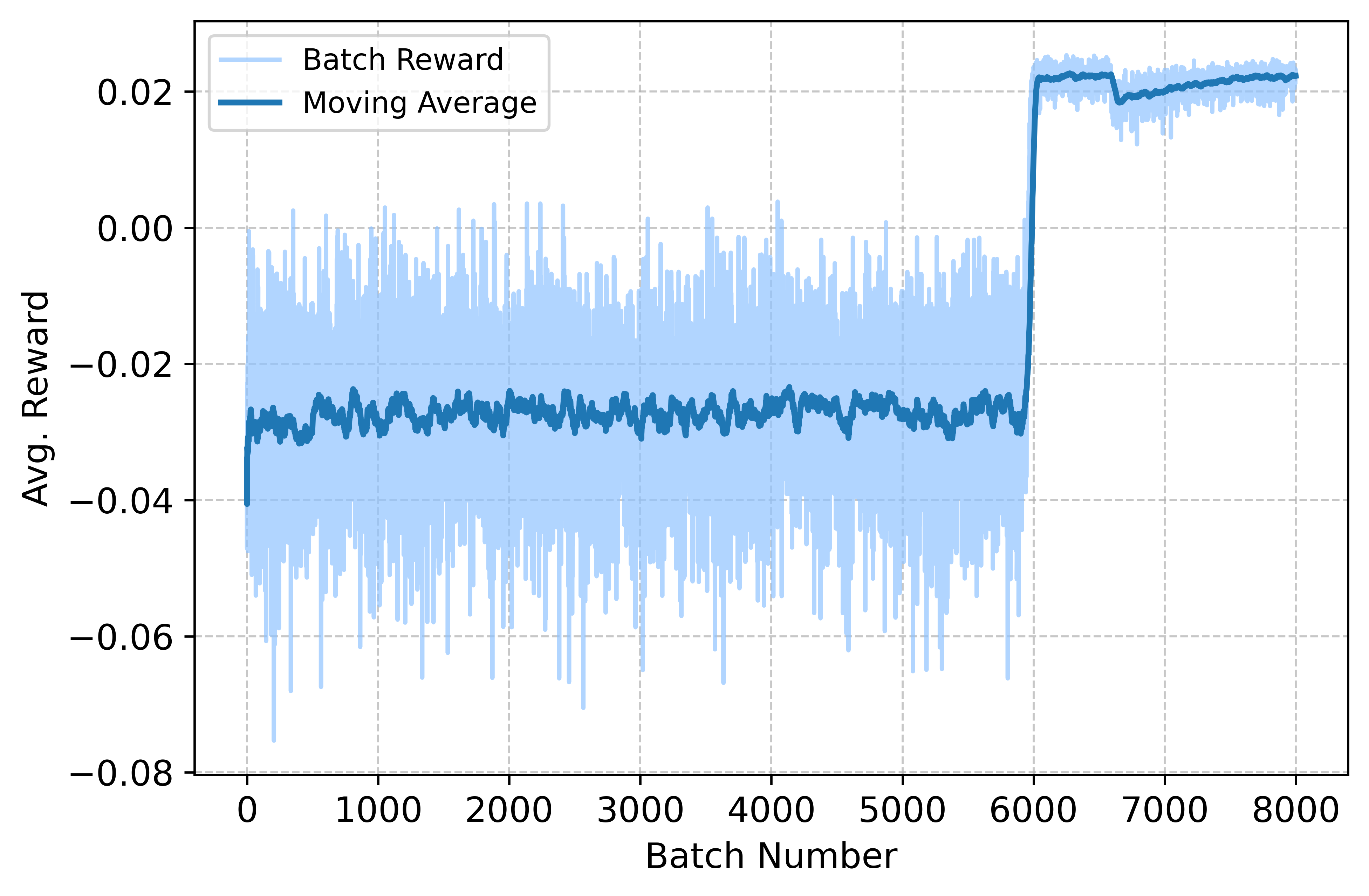}
    \caption{Convergence plot for optimal $n_r^\star$.}
    \label{fig:conv_plot_opt}
\end{figure}}

\subsection{Empirical Policy Performance Evaluation}\label{sec:empir_res}
In this section, we evaluate the performance of a trained policy $\pi^{\star}_{\boldsymbol{\phi}}$ which is the output of Alg.~\ref{algo:reinforce} post convergence, and compare its performance to that of a baseline cut-off policy $\pi^b$, defined as
\begin{equation}
    \pi^b(s,a, k) = \begin{cases}
        \text{move} & P_C(s,k) \geq P_{C,max} \text{ and } k = K-3, \\
        \text{stay} & \text{o.w.} 
    \end{cases}
\end{equation}
Each policy is evaluated via $M = 1000$ Monte Carlo (MC) trials using both synthetic and historical datasets. The policy performance via three metrics: accurate maneuvering in high risk conjunctions, total propellant consumption, and average propellant consumption. 

\textbf{Maneuvering Accuracy}. We use confusion matrices to contrast each policy's maneuvering accuracy for high vs low collision risks.  Each element of the confusion matrix $\boldsymbol{C} \in\reals^{2\times2}$ denotes the proportion between the realized collision risk (high vs low) and the maneuver executed (move or stay). Each matrix row corresponds to a high (true) vs low (false) collision risk, and each column corresponds to a move (positive) vs stay (negative) maneuver. 
For a policy $\pi$ evaluated under $M$ MC runs, let $c_{ij}, i \in \{T,F\}, j \in \{P, N\}$ denote the empirical count of the outcome $(i,j)$. The empirical confusion matrix is given by
\begin{equation}\label{eqn:def_confusion}
   \textstyle \boldsymbol{C}(\pi) = 
    \begin{bmatrix}
        C_{TP} &  C_{FN} \\ C_{FP} & C_{TN}
    \end{bmatrix}, \, C_{ij}= \frac{1}{M}c_{ij}/\sum_{\hat{j}} c_{i\hat{j}}, \quad \forall (i,j) \in  \{T,F\}\times \{P, N\}. 
\end{equation}
In addition, we compute the accuracy of each confusion matrix via $A(\boldsymbol{C})$~\citep{powers2020}, given by  
\begin{equation}\label{eqn:acc}
    A(\boldsymbol{C}) = \frac{\sum_iC_{ii}}{\sum_{i,j}C_{ij}}.
\end{equation}
The accuracy metric ranges from $A(\boldsymbol{C})=1$ ($\boldsymbol{C}$ is diagonal) to $A(\boldsymbol{C})=0$ ($\boldsymbol{C}$ is hollow). The ideal maneuver guidance policy is diagonal: all high risk events are maneuvered and all low risk events are not maneuvered.  

\textbf{Propellant Consumption}. We evaluate each policy's performance by comparing their propellant consumption efficiency. The total propellant consumption $m_{p, tot} \in \reals$ is the empirical sum of propellant consumed (Eq.~\eqref{eqn:fuel}) across $M = 1000$ MC trials, and the average propellant consumption is the average over all trials. Let $m_p^i \in \reals$ denote MC trial $i$'s propellant consumption, then the total and average propellant consumption are given by 
\begin{equation}\label{eqn:tot_avg_mp}
    \textstyle m_{p,tot} = \sum_{i=1}^M m_p^i, \quad m_{p,avg} = \frac{1}{M}\sum_{i=1}^M m_p^i.
\end{equation}


\textbf{Default Settings: Policy Performance}.
In this section, we compare the policy performance of the trained policy $\pi_{\boldsymbol{\phi}}^\star$ to a baseline cut-off policy $\pi^b$ for \textbf{Default} parameters in Tab.~\ref{tab:case_params} using both synthetic and historical CDM data. We evaluate both policies over $M = 1000$ MC trials in each dataset.

\begin{figure}[h!]
    \centering
    \includegraphics[width=\linewidth]{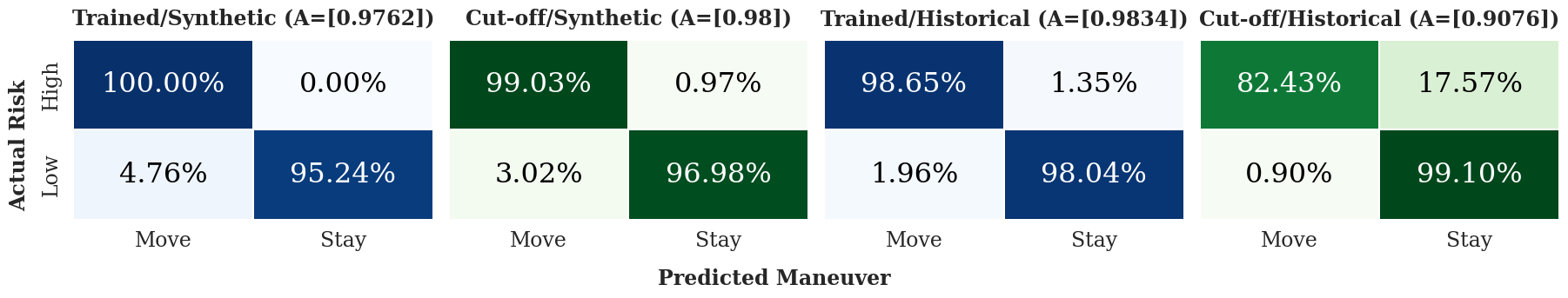}
    \caption{Confusion matrices for trained (blue) $\pi_{\boldsymbol{\phi}}^\star$ and cut-off (green) $\pi_{\boldsymbol{\phi}}^b$ strategies along with respective accuracies $A$ from Eq.~\eqref{eqn:acc} under default settings with synthetic and Historical CDMs.}
    \label{fig:default_confusion}
\end{figure}
Fig.~\ref{fig:default_confusion} visualizes the  confusion matrix results. Across both synthetic and historical CDM datasets, we observe that the trained policy $\pi_{\boldsymbol{\phi}}^\star$ achieves the highest move rate for high risk conjunction events (first from left), while the cut-off policy achieves the lowest move rate for high risk conjunction events (last from left). Additionally, the trained policy $\pi_{\boldsymbol{\phi}}^\star$ outperforms cut-off policy $\pi^b$ over both synthetic CDMs as well as historical CDMs. 

\begin{figure}[h!]
    \begin{minipage}{0.45\textwidth}
        \centering
        \begin{table}[H]
        \centering
        \begin{tabular}{l c c c}
            \toprule
            \textbf{Policy} & \textbf{Total} & \textbf{Avg./CAM} & \textbf{No. of Man.}  \\ 
            \midrule
            Trained  & $\approx$ \textbf{0.93 kg} & $\approx$  \textbf{3.57 g} & 260\\ 
            Cut-Off     & $\approx$ 1.08 kg & $\approx$ 4.72 g & 228 \\ 
            \bottomrule
        \end{tabular}
        \caption{Propellant consumption comparison for synthetic CDMs under default model parameters.}
        \label{tab:fuelusage}
        \end{table}
    \end{minipage}
    \hfill
    \begin{minipage}{0.45\textwidth}
        \centering
        \begin{table}[H]
        \centering
        \begin{tabular}{l c c c}
            \toprule
            \textbf{Policy} & \textbf{Total} & \textbf{Avg./CAM} & \textbf{No. of Man.} \\ 
            \midrule
            Trained  & $\approx$ 0.75 kg & $\approx$  \textbf{3.44 g} & 218 \\ 
            Cut-Off     & $\approx$ \textbf{0.61 kg} & $\approx$ 4.69 g & 129 \\ 
            \bottomrule
        \end{tabular}
        \caption{Propellant consumption comparison for historical CDMs under default model parameters.}
        \label{tab:fuelusage_cdms}
        \end{table}
    \end{minipage}
\end{figure}
Tab.~\ref{tab:fuelusage} and~\ref{tab:fuelusage_cdms} show the propellant consumption efficiency results using the total and average propellant consumption in Eq.~\eqref{eqn:tot_avg_mp}, as well as the number of maneuvers as comparison metrics. 
Over synthetic CDM dataset, the trained policy $\pi_{\boldsymbol{\phi}}^\star$ consumes less propellant than the cut-off policy $\pi^b$ both in total and on average per CAM, but initiates more maneuvers. Over historical CDM dataset, the trained policy $\pi_{\boldsymbol{\phi}}^\star$ consumes less propellant per CAM but consumes more propellant overall when compared to the fuel consumption of the cut-off policy $\pi^b$, this is partially due to $\pi_{\boldsymbol{\phi}}^\star$ initiating almost twice as many maneuvers as $\pi^b$. 
We conclude that for both the synthetic and historical CDMs, the trained policy $\pi_{\boldsymbol{\phi}}^\star$ is more conservative and thus captures more high risk events at the expense of a higher false-positive rate, as well as total propellant consumed for historical CDMs. These results may also be due to a lack of high risk conjunction events in the historical CDM dataset, which may skew the evaluation outcome.

\textbf{Variation 1: Policy Performance under Hard Body and Phase Variations}.
In this section, we explore how varying $R_{HB}$ and $\Delta\theta$ influence the trained policy $\pi_{\boldsymbol{\phi}}^\star$ and the baseline cut-off policy $\pi_{\boldsymbol{\phi}}^b$'s performance under 
\textbf{Var. 1 ($R_{HB}/\Delta\theta$)} model parameters from Tab.~\ref{tab:case_params} and~\ref{tab:binary_options}. We observed that since parameter set with index j=1 in Tab.~\ref{tab:binary_options} did not converge (DNC),  we do not evaluate this parameter combination here. 

\begin{figure}[h!]
    \centering
    \includegraphics[width=\linewidth]{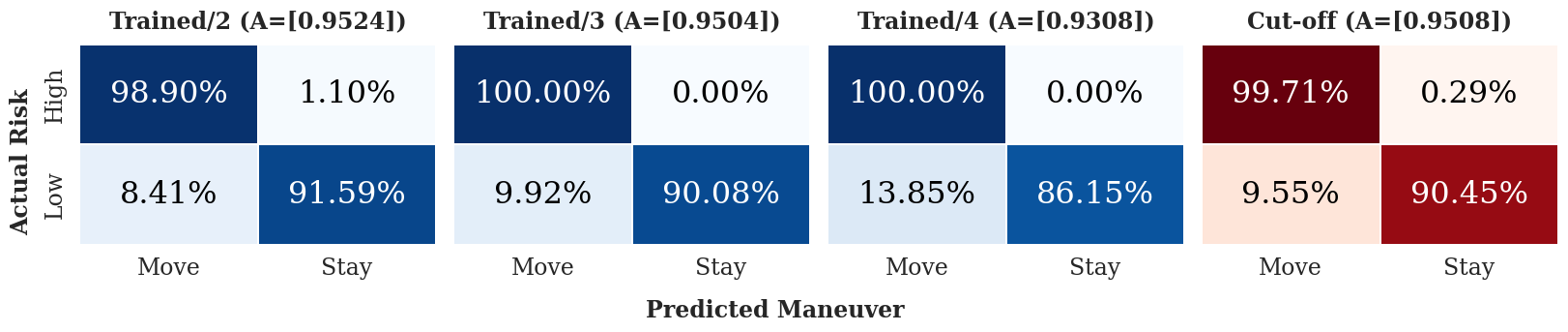}
    \caption{Confusion matrices for HBR and phase combinations (blue) $\pi_{\boldsymbol{\phi}}^j,\,\forall j\in\{2,3,4\}$ (Index 1 DNC, not included) and cut-off (red) $\pi^b$ strategies along with respective accuracies $A$ from Eq.~\eqref{eqn:acc}  under synthetic CDMs.}
    \label{fig:var1_actions_synth}
\end{figure}
Fig.~\ref{fig:var1_actions_synth} visualizes the confusion matrix results for synthetic CDMs.  
The trained policy under all parameter sets achieve comparable accuracy to the cut-off policy. 
The parameter set with index $j=2$ (variable $R_{HB}$,  fixed $\Delta\theta$) has the highest accuracy, approximately $0.2\%$ greater than that of index $j=4$ (fixed $R_{HB}$, fixed $\Delta\theta$) despite slower convergence rate in Fig.~\ref{fig:convergence_binary}. 
Furthermore, the trained policy under index $j=4$ has both the smallest total and average propellant consumption, while still having a smaller total number of maneuvers compared to the cut-off policy.

\begin{figure}[h!]
    \centering
    \includegraphics[width=\linewidth]{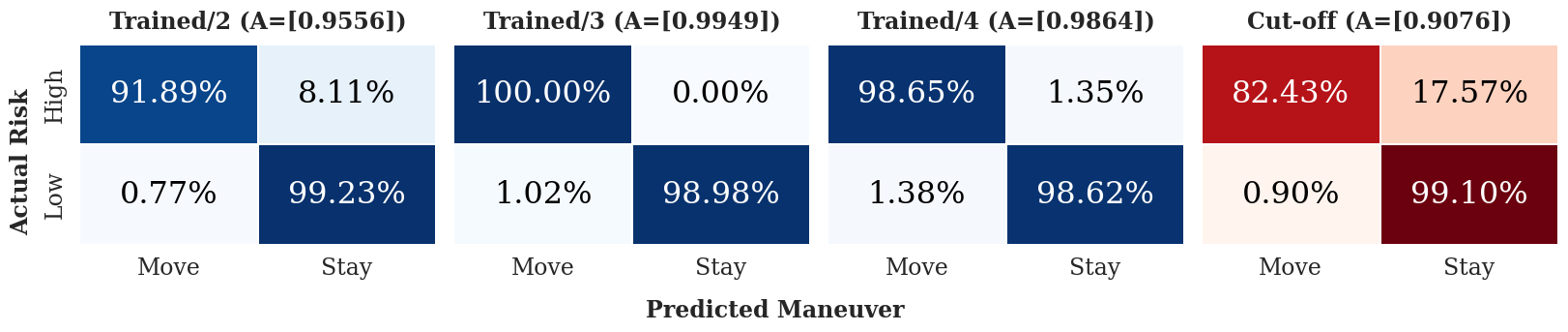}
    \caption{Confusion matrices for HBR and phase combinations (blue) $\pi_{\boldsymbol{\phi}}^j,\,\forall j\in\{2,3,4\}$ (Index 1 DNC, not included) and cut-off (red) $\pi^b$ strategies along with respective accuracies $A$ from Eq.~\eqref{eqn:acc}  under historical CDMs.}
    \label{fig:var1_actions_hist}
\end{figure}

Fig.~\ref{fig:var1_actions_hist} visualizes the confusion matrix results for historical CDMs. The trained policy under all the parameter sets presented achieve comparable accuracy to the cut-off policy. In this case, the parameter set with index 3 (fixed $R_{HB}$, variable $\Delta\theta$) results in the highest accuracy, about 0.7\% above that of index 4 (fixed $R_{HB}$, fixed $\Delta\theta$). 

Furthermore, Tab.~\ref{tab:fuelusage_var} and~\ref{tab:fuelusage_cdms_var} show the propellant consumption efficiency results for all $j=\{1,2,3,4\}$ parameter sets in Tab.~\ref{tab:binary_options}. We observe in Tab.~\ref{tab:fuelusage_var} that the trained policy under parameter set $4$ has the best performance in both total and average fuel consumption, while initiating the least number of maneuvers. Finally, we observe in Tab.~\ref{tab:fuelusage_cdms_var} that, even though the cut-off policy has the lowest total propellant consumed for historical CDMs, the trained policy under parameter set $4$ has the smallest average propellant consumption per CAM as it has the highest number of maneuvers.
\begin{figure}[h!]
    \begin{minipage}{0.48\textwidth}
        \centering
        \begin{table}[H]
        \centering
        \begin{tabular}{l c c c}
            \toprule
            \textbf{Policy/$j$} & \textbf{Total} & \textbf{Avg./CAM} & \textbf{No. of Man.}  \\ 
            \midrule
            Trained/1  & \textit{DNC} & \textit{DNC} & \textit{DNC}\\ 
            Trained/2  & $\approx$ 1.29 kg & $\approx$  6.17 g & 209\\
            Trained/3  & $\approx$ 1.06 kg & $\approx$  4.49 g & 236\\
            \textbf{Trained/4}  & $\approx$ \textbf{0.89 kg} & $\approx$  \textbf{3.98 g} & 224\\
            Cut-Off     & $\approx$ 1.08 kg & $\approx$ 4.72 g & 228 \\ 
            \bottomrule
        \end{tabular}
        \caption{Propellant consumption comparison for synthetic CDMs under HBR and phase variations.}
        \label{tab:fuelusage_var}
        \end{table}
    \end{minipage}
    \hfill
    \begin{minipage}{0.48\textwidth}
        \centering
        \begin{table}[H]
        \centering
        \begin{tabular}{l c c c}
            \toprule
            \textbf{Policy/$j$} & \textbf{Total} & \textbf{Avg./CAM} & \textbf{No. of Man.}  \\ 
            \midrule
            Trained/1  & \textit{DNC} & \textit{DNC} & \textit{DNC}\\
            Trained/2  & $\approx$ 0.71 kg & $\approx$  5.69 g & 124\\
            Trained/3  & $\approx$ 0.73 kg & $\approx$  4.85 g & 150\\
            Trained/4  & $\approx$ 0.74 kg & $\approx$  \textbf{4.21 g} & 176\\
            Cut-Off     & $\approx$ \textbf{0.61 kg} & $\approx$ 4.69 g & 129 \\ 
            \bottomrule
        \end{tabular}
        \caption{Propellant consumption comparison for historical CDMs under HBR and phase variations.}
        \label{tab:fuelusage_cdms_var}
        \end{table}
    \end{minipage}
\end{figure}

\textbf{Variation 2: Policy Performance under Propellant-risk Trade-off Variations}.
In this study, we explore the effects of varying reward weight parameter $\eta \in \reals$ on the policy performance for the trained policy $\pi_{\boldsymbol{\phi}}^\eta$ and the cut-off policy $\pi^b$ under \textbf{Var. 2 ($\eta$)} model parameters from Tab.~\ref{tab:case_params}. 

\begin{figure}[h!]
    \centering
    \includegraphics[width=0.8\linewidth]{fig/eta_percent_action_distrib_combined.png}
    \caption{Confusion matrix values $\boldsymbol{C}_{ij}^{q}=\boldsymbol{C}_{(i,j)}(\pi_{\boldsymbol{\phi}}^q),\,\forall q\in\{\eta,b\}$ (Eq.~\eqref{eqn:def_confusion}) under variations in $\eta\in[0,1]$ (Eq.~\eqref{eqn:reward})  for trained policies ($\eta$) and cut-off policy ($b$) respectively under synthetic CDMs.}
    \label{fig:eta_confusion_matrix}
\end{figure}
Fig.~\ref{fig:eta_confusion_matrix} visualizes confusion matrix changes under increasing $\eta$ for both synthetic and historical CDMs by sampling $\eta\in[0,1]$ from 0 to 1 at intervals of $0.1$, which results in a region $\eta\in[0,0.6]$ where $C_{ij}^\eta = \boldsymbol{C}_{(i,j)}(\pi_{\boldsymbol{\phi}}^\eta)$ (Eq.~\eqref{eqn:def_confusion}) remains unchanging and stable. We denote $\eta\in[0,0.6]$ the ``stable-$\eta$" region. Within this region, we notice that the maneuver distribution is more stable for the synthetic CDMs than for the historical CDMs. 
The stability within the ``stable-$\eta$" region can be related to the reward function Eq.~\eqref{eqn:reward}, where  $\eta<0.6$ corresponds to putting more importance on collision risk minimization over propellant consumption minimization. In Fig.~\ref{fig:eta_propellant} we observe, for $\eta\leq 0.6$, that the trained policy consumes more propellant in total and on average than the cut-off policy on the synthetic CDM dataset, despite having near identical maneuver initiations. As $\eta$ increases to $\eta=0.6$ (still within the ``stable-$\eta$" region), the total and average propellant consumption for the trained policy begin to oscillate about the cut-off value over both datasets, and then begin to decrease for $\eta>0.6$. Additionally, within the ``stable-$\eta$" region, the number of maneuvers for the trained policy under synthetic CDMs remains relatively constant, but begin to increase as $\eta>0.6$. We conclude that, when $0\leq\eta\leq 0.6$, the trained policy has a greater accuracy and stable action distribution at the expense of a higher and unstable total and average propellant consumption. However, as $\eta$ increases past $0.6$, the trained policy will focus on minimizing propellant consumption at the expense of decreasing action distribution accuracy, where we observe an increase in false-positive rates and a decrease in true-negative rates. Thus, we recommend to set $\eta=0.6$ to obtain a trained policy that maximizes true-positive and true-negative rates (maximizing accuracy) while minimizing propellant consumption simultaneously.


\begin{figure}[h!]
    \centering
    \includegraphics[width=\linewidth]{fig/eta_propellant_combined_tripple.png}
    \caption{Effect of variation in $\eta\in[0,1]$ (Eq.~\eqref{eqn:reward}) on total propellant and average propellant per CAM (Eq.~\eqref{eqn:tot_avg_mp}) by strategy for synthetic and historical CDMs ($\eta=1$ DNC).}
    \label{fig:eta_propellant}
\end{figure}

\textbf{Significance to Astrodynamics and Space-flight Mechanics}. Our results demonstrate that MDP-based maneuver guidance are a feasible approach for automating and improving the propellant efficiency of the CAM guidance process. This is critical when considering the increasing number of conjunction events and the expanding mega-constellations residing in LEO. For example, SpaceX's Starlink satellites claimed to perform about 27 maneuvers per satellite in 2024~\citep{SpaceX2024SpaceReport}. Clearly, if we evaluate the total consumption of thousands of satellites, halving the propellant mass per maneuver would be very beneficial, both because it would extend the satellite's lifetime and it would save on operation and propellant costs. 
\section{Conclusion}
We formulate a continuous-state, finite-horizon MDP with a discrete action space and stochastic transition dynamics to improve propellant efficiency in CAMs. Particular attention was paid to the dynamic evolution of critical conjunction parameters and the propellant usage decrease for CAMs that are initiated prior to cut-off time. We explored how varying the parameters and hyper-parameters related to the modeling and training influenced the RL-PG training complexity and its performance on an empirical level. The results for each case study carried out are promising for using both synthetic and historical CDM data to augment existing CAM guidance.

The results we presented might serve insightful for future advancements in autonomous satellite collision avoidance (CA) decision-making. We outline here some key areas where advancements could be made in relation to this work. The accuracy of the model could be improved upon by introducing more state variables, such as the combined-covariance matrix, the relative position, and the relative velocity. Their inclusion could result in a more realistic collision dynamics. With respect to realism, the model could be further improved by introducing gravitational perturbation ($J_2$-effect) and drag, which could help simulate a more accurate environment. Furthermore, by generating more high risk CDM data, the RL-PG training complexity and performance could be improved, and the relation between the action distribution and the quantity of high-risk data available could be explored.

\bibliographystyle{aasjournal}
\bibliography{sample7}{}

\end{document}